\documentclass[lettersize,journal]{IEEEtran}
\usepackage{cite}
\usepackage{amsmath,amssymb,amsfonts}
\usepackage{algorithmicx}
\usepackage{algcompatible}
\usepackage{algorithm}
\usepackage[noend]{algpseudocode}
\usepackage{graphicx}
\usepackage{textcomp}
\usepackage{multirow}
\usepackage{booktabs, tabularx}
\usepackage{amsmath}
\usepackage{booktabs}
\usepackage{subcaption} 
\usepackage{float} 
\usepackage{tabularx}    
\usepackage{multirow}    
\usepackage{threeparttable} 

\usepackage{picinpar}
\usepackage{amsmath}
\usepackage{url}
\usepackage{flushend}
\usepackage[latin1]{inputenc}
\usepackage{colortbl}
\usepackage{soul}
\usepackage{pifont}
\usepackage{color}
\usepackage{alltt}
\usepackage[hidelinks]{hyperref}
\usepackage{enumerate}
\usepackage{siunitx}
\usepackage{breakurl}
\usepackage{epstopdf}
\usepackage{pbox}

\usepackage{textcomp}
\usepackage{stfloats}
\usepackage{verbatim}
\begin{document}
\title{MMH-Planner: Multi-Mode Hybrid Trajectory Planning Method for UAV Efficient Flight Based on Real-Time Spatial Awareness}

\author{
	\vskip 1em
	
	Yinghao Zhao, Chenguang Dai, Liang Lyu, Zhenchao Zhang, Chaozhen Lan, and Hong Xie

	\thanks{
	
		Yinghao Zhao, Chenguang Dai, Liang Lyu, Zhenchao Zhang, and Chaozhen Lan are with the School of Surveying and Mapping, Information Engineering University, Zhengzhou, 450001, China (e-mail: zhaoyinghao@whu.edu.cn; cgdai2008@163.com; lvliangvip@163.com; zhzhc\_1@163.com; lan\_cz@163.com). 
		
		Hong Xie is with the School of Geodesy and Geomatics, Hubei Luojia Laboratory, Wuhan University, Wuhan 430079, China (e-mail: hxie@sgg.whu.edu.cn).
	}
}

\markboth{Journal of \LaTeX\ Class Files,~Vol.~14, No.~8, August~2021}%
{Shell \MakeLowercase{\textit{et al.}}: A Sample Article Using IEEEtran.cls for IEEE Journals}

\IEEEpubid{0000--0000/00\$00.00~\copyright~2021 IEEE}


\maketitle

\begin{abstract}
Motion planning is a critical component of intelligent unmanned systems, enabling their complex autonomous operations. However, current planning algorithms still face limitations in planning efficiency due to inflexible strategies and weak adaptability. To address this, this paper proposes a multi-mode hybrid trajectory planning method for UAVs based on real-time environmental awareness, which dynamically selects the optimal planning model for high-quality trajectory generation in response to environmental changes. First, we introduce a goal-oriented spatial awareness method that rapidly assesses flight safety in the upcoming environments. Second, a multi-mode hybrid trajectory planning mechanism is proposed, which can enhance the planning efficiency by selecting the optimal planning model for trajectory generation based on prior spatial awareness. Finally, we design a lazy replanning strategy that triggers replanning only when necessary to reduce computational resource consumption while maintaining flight quality. To validate the performance of the proposed method, we conducted comprehensive comparative experiments in simulation environments. Results demonstrate that our approach outperforms existing state-of-the-art (SOTA) algorithms across multiple metrics, achieving the best performance particularly in terms of the average number of planning iterations and computational cost per iteration. Furthermore, the effectiveness of our approach is further verified through real-world flight experiments integrated with a self-developed intelligent UAV platform.
\end{abstract}



\definecolor{limegreen}{rgb}{0.2, 0.8, 0.2}
\definecolor{forestgreen}{rgb}{0.13, 0.55, 0.13}
\definecolor{greenhtml}{rgb}{0.0, 0.5, 0.0}

\section{Introduction}
\label{sec:introduction}
\IEEEPARstart{W}{ith} the continuous advancement of intelligent technologies, unmanned systems have emerged as a focus of global attention\cite{guo2025global,du2025efficient,dominguez2025cooperative,zhou2026hfch,wang2025unlocking}. Owing to their unique advantages, such as high flexibility and strong adaptability, unmanned aerial vehicles (UAVs) have been widely deployed across numerous fields\cite{zhang2025hybrid,liu2024dawn,zhao2023autonomous,zhou2023racer}. However, in most current practical applications, UAV operations still rely on human control or pre-programmed flight paths, which falls short of meeting the demands for safe, autonomous, intelligent, and efficient operations, especially in unknown and complex environments such as beneath forest canopies, indoors, or underground. Consequently, there is a pressing need to enhance the autonomy and intelligence of UAVs.
\vspace{-1mm}
\subsection{Existing Problems}
As a critical component of intelligent UAVs, the motion planning module generates real-time, high-quality trajectories that are safe, smooth, and dynamically feasible\cite{zhou2022swarm,zhao2023robust}. This is achieved by integrating real-time scene perception data with target location information, ensuring the UAV reaches its destination securely, stably, and rapidly. Despite significant advancements in recent years\cite{zhou2019robust,zhou2020ego,tordesillas2021mader}, persistent challenges persist in motion planning algorithms. First, current motion planning algorithms fail to incorporate the complexity of real-time navigational environments as prior information, resulting in suboptimal rationality and adaptability of the optimization models. Second, existing algorithms primarily fall into two categories: hard-constraint optimization and soft-constraint optimization. Hard-constraint algorithms deliver high planning quality but suffer from low computational efficiency and adaptability. Soft-constraint algorithms, while highly efficient and adaptable, struggle to strictly satisfy all constraints in their planning results, preventing an optimal balance between quality and efficiency. 
\IEEEpubidadjcol
Furthermore, current algorithms often adopt small-horizon, high-frequency replanning strategies to enhance trajectory optimality and safety. However, this strategy generates excessive unnecessary replanning (no or limited improvement in terms of security, continuity, and optimality), wasting limited computational resources. In summary, how to achieve rational, efficient, and high-quality trajectory generation is still an open question. 
\vspace{-1mm}
\subsection{Contributions}
To address the aforementioned challenges, this paper proposes a hybrid intelligent replanning algorithm capable of autonomously selecting the optimal planning model based on real-time navigation spatial awareness, named MMH (Multi-Mode Hybrid)-Planner. First, we design a goal-oriented spatial awareness method that rapidly assesses the impact of obstacles within the upcoming area using goal-oriented sensing ray, providing a priori guidance for subsequent trajectory planning. Second, a hybrid intelligent planning mechanism incorporating three distinct modes (Fast, Standard, and Emergency) is developed based on the characteristics of different planning models. This mechanism fully integrates the advantages of various trajectory optimization models, enabling real-time selection of the most appropriate planning mode according to the awareness of the upcoming area. This ensures maximum fulfillment of trajectory generation efficiency, quality, and flight safety. Finally, a lazy re-planning mechanism is designed to trigger trajectory replanning only when necessary (speed or safety deteriorates). This approach minimizes the waste of onboard computational resources while maintaining flight safety and optimality. The main contributions of this work are summarized as follows:

\begin{enumerate}[1)] 
    \item A goal-oriented spatial awareness method is introduced for rapid safety assessment of the upcoming environments, providing effective guidance for trajectory planning.
    \item A multi-mode hybrid replanning mechanism is proposed, which flexibly selects the optimal planning model for trajectory generation based on the spatial awareness. This mechanism fully leverages the strengths of three unique planning models to enhance the replanning efficiency.
    \item A lazy replanning strategy is designed to reduce unnecessary replanning iterations and conserve limited onboard computational resources. 
    \item Sufficient comparison experiments are conducted in simulation, and real-world experiment is also carried out to validate the effectiveness of the proposed method.
\end{enumerate}

\begin{figure*}[!ht]
\centering
\includegraphics[width=\textwidth]{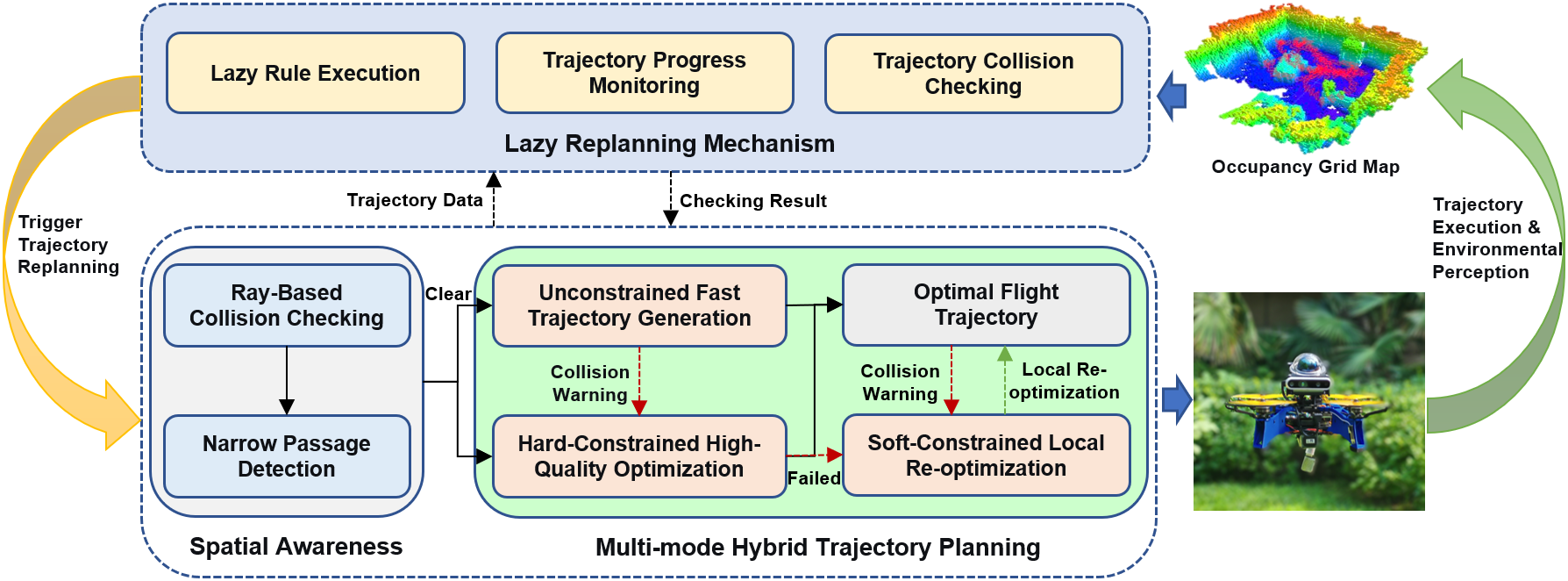}
\caption{Framework of the multi-mode hybrid planner.}
\label{fig1}
\vspace{-3mm}
\end{figure*}
\section{Releated Work}
In recent years, a variety of motion planning algorithms have been proposed and adopted. Based on their underlying principles, widely used high-quality motion planning algorithms can be broadly categorized into two main classes: soft-constrained optimization\cite{zhao2021robust,ratliff2009chomp,usenko2017real} and hard-constrained optimization\cite{mellinger2011minimum,richter2016polynomial,tordesillas2019faster}. 
\vspace{-1mm}
\subsection{Hard-constrained Optimization}
Hard-constrained optimization methods originated with the minimum snap trajectory generation approach introduced by Mellinger and Kumar\cite{mellinger2011minimum}, which represents UAV trajectories using piecewise polynomials and solves a quadratic programming (QP) problem. Later, to enhance trajectory safety, many researchers have adopted a two-stage trajectory generation method\cite{wang2022geometrically,tordesillas2019faster,ren2022bubble}: first constructing a safe flight corridor (SFC) (composed of cubes, spheres, or convex polyhedra) to represent safe space, and then solving a QP problem within this space to generate a safe flight trajectory. To efficiently compute spatiotemporally optimal trajectories, \cite{wang2022geometrically} proposed a novel MINCO trajectory representation method, enabling joint spatial-temporal optimization and significantly improving trajectory quality. Building on MINCO, \cite{ren2022bubble} introduced the Bubble Planner, which further enhanced planning efficiency through innovations in SFC generation and reuse. Subsequently, the same team proposed an unmanned aerial system called SUPER\cite{ren2025safety}, capable of autonomous obstacle avoidance at speeds up to 20 m/s in unknown complex environments. However, this method essentially follows the SFC-based hard-constrained dual-trajectory planning strategy of Faster\cite{tordesillas2019faster}. While it ensures high trajectory quality, its planning efficiency remains relatively low. Overall, hard-constrained methods produce high-quality trajectories, but their need to construct SFCs and strictly satisfy constraints leads to relatively low computational efficiency and susceptibility to sensor noise. 
\vspace{-1mm}
\subsection{Soft-constrained Optimization}
In contrast, soft-constrained optimization methods offer better real-time performance and adaptability. They formulate the path planning problem as a nonlinear optimization problem by incorporating safety, smoothness, and dynamic feasibility into the cost function. \cite{zhou2019robust} employed a hybrid A* search to generate an initial kinodynamic path and efficiently optimized the trajectory using B-splines. However, this method remains prone to local minima and planning failure when the initial trajectory quality is poor. To address this, the authors later significantly improved planning success rates using topological path guidance\cite{zhou2021raptor}. To eliminate the dependency on computationally expensive euclidean signed distance fields (ESDF), \cite{ye2020tgk} proposed TGK-Planner, which uses an improved kinodynamic RRT* method for initial trajectory search and trajectory optimization. \cite{zhou2020ego} also introduced the well-known EGO-Planner, which directly computes distance gradients from collision-free geometric paths instead of relying on ESDF, substantially improving computational efficiency. Later, \cite{quan2021eva} integrated real-time environmental information into planning decisions, enabling UAVs to dynamically adjust flight aggressiveness, thereby enhancing adaptability and planning performance. However, although soft-constrained methods perform well in real-time performance and adaptability, their elastic optimization nature makes it difficult to guarantee high-quality results or strictly satisfy all predefined constraints.

In summary, both hard-constrained methods (strictly satisfy constraints) and more flexible soft-constrained methods suffer from their respective drawbacks, making it challenging for them to plan flexibly and efficiently in complex scenarios. 

\section{Technical Approach}
\subsection{System Overview}
The main workflow of this method is illustrated in Fig. \ref{fig1}, comprising three key components: goal-oriented spatial awareness of the navigation environment, multi-mode hybrid trajectory planning mechanism, and lazy replanning strategy. First, to rapidly assess the impact of obstacles within the upcoming navigation environment on flight tasks, this paper utilizes the maximum sensing range of sensors to check obstacles in the target direction and adjacent areas. Based on these checking results, it determines whether obstacles in the upcoming spaces affect the current flight. Second, based on the navigation spatial awareness results and the characteristics of different trajectory optimization models, the optimal trajectory optimization mode is specifically selected to achieve the best balance between safety, quality, and efficiency. Subsequently, the designed lazy replanning strategy triggers trajectory replanning when necessary to ensure flight safety and continuity. Finally, as the UAV moves, data acquired by sensors continuously updates the local map in real time, providing essential data for avoidance planning. 
\vspace{-1mm}
\subsection{Goal-oriented Spatial Awareness}
\label{GEA}

In traditional planning methods, the trajectory generation process primarily relies on obstacle positions for collision-free path search and optimization, without leveraging the spatial distribution of obstacles to adjust the planning strategy. This causes the limited flexibility, making it difficult to plan efficiently in complex environments. To address this limitation, unlike conventional approaches that directly perform trajectory search and optimization, this paper introduces a goal-oriented environment complexity awareness method. This method evaluates the obstacle interference between the UAV's current position and the target position, thereby providing efficient and effective information support for the subsequent flexible selection of the optimal trajectory planning algorithm. The main procedure of the method is outlined in Algorithm 1.

\begin{figure}[!ht]
\centering
\includegraphics[width=\linewidth]{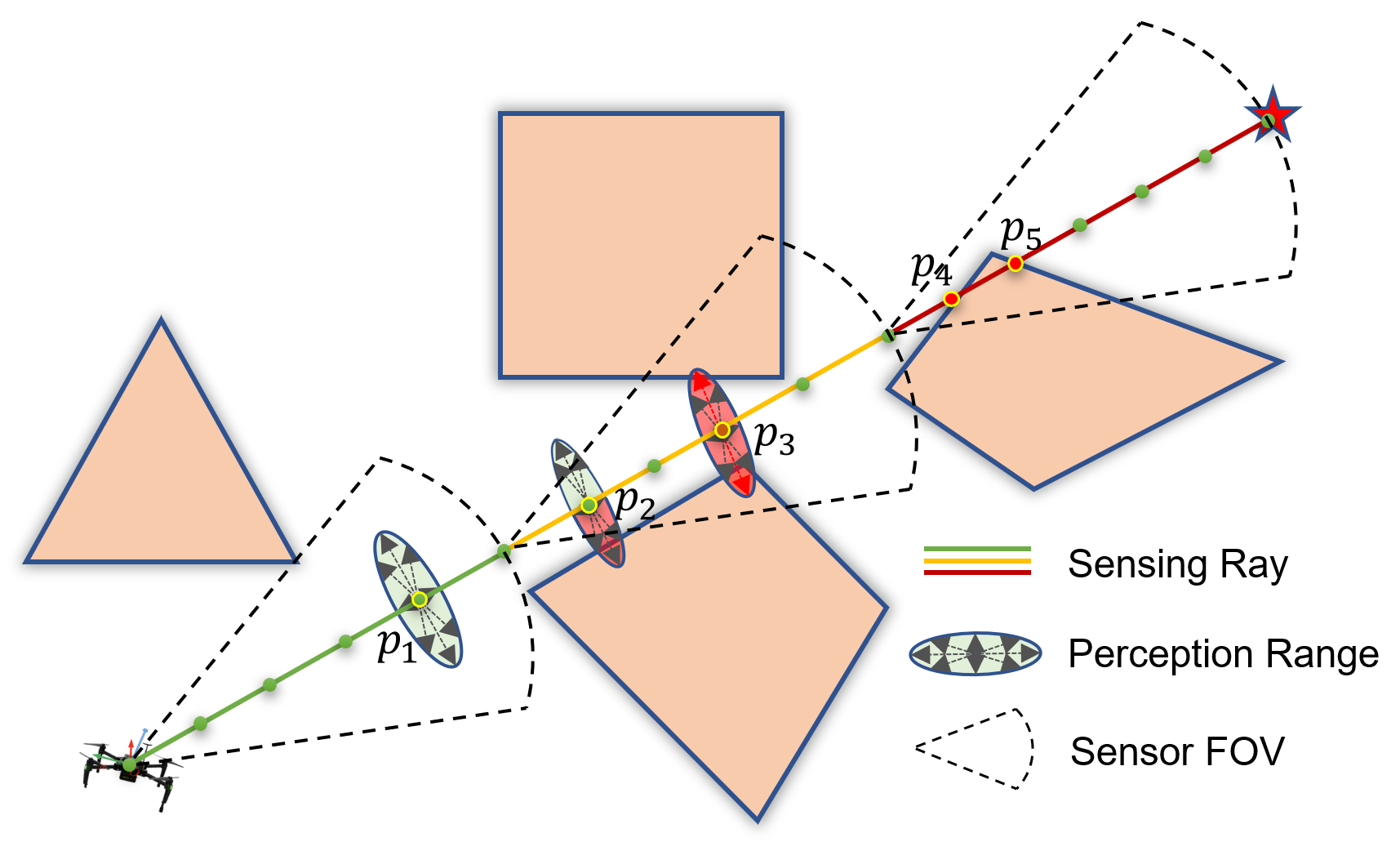}
\caption{Schematic diagram of the goal-oriented spatial awareness method. This figure illustrates three distinct perception scenarios. Specifically, the regions corresponding to green rays indicate no obstacles are detected, representing a safe environment. The areas covered by yellow rays reveal passage through narrow regions via local perception, while red rays denote scenarios where obstacle checking confirms that the rays pass through obstacles. Both the latter two cases involve obstacles that compromise the current flight safety, thereby necessitating obstacle avoidance planning.}
\label{fig2}
\vspace{-4mm}
\end{figure}

First, based on the sensor's maximum perception range $r$, the endpoint position $G_M$ of the sensing ray is computed using the UAV's current position $P$ and the goal position $G$. The sensing ray $\overrightarrow{PG_M}$ (shown as the colored segment in Fig. \ref{fig2}) is then uniformly sampled to obtain a sequence of sampling points $p_{\text{list}}\{p_0,\cdots,p_n\}$ (colored dots in Fig. \ref{fig2}), with the sampling interval determined by the resolution of the occupancy grid map (Lines 1-2). Subsequently, safety inspection is performed based on the sampling results. If a sampling point $p_n$ lies inside an obstacle (e.g., points $p_4$ and $p_5$ in Fig. \ref{fig2}), the UAV's flight safety is considered compromised, and the safety flag $S$ is set to False (Lines 3-6). Otherwise, narrow space detection is conducted. The narrow space detection method adopted in this paper is computationally simple yet efficient. It is primarily based on the normal directions to the vector between adjacent discrete points, performing bidirectional obstacle detection within a certain distance along these normals. If obstacles are detected on both sides of the normal (as at point $p_3$ in Fig. \ref{fig2}), the current traversable area is considered to contain a narrow passage, and the flight safety is also regarded as compromised, with $S$ also marked as False (Lines 7-9). The awareness result will also directly determine the selection of the planning model in the next part.
\begin{algorithm}[t]
\caption{Goal-Oriented Complexity Awareness Algorithm for Passage Environment}
\label{alg:passage_env}
\begin{algorithmic}[1]
\Require UAV's position $P$, goal's position $G$, maximum sensor range $r$ and map data $M$
\Ensure Passage environment safety $S$ (True or False)
\State $P_g \gets \text{SenseRay}(P, G, r)$
\State $p_{list} \gets P_g.\text{discrete}()$
\State $n \gets 1$
\While{$0 < n < p_{list}.\text{size}() - 1 $}
    \State $S \gets \text{JudgePointSafe}(p_{n-1}, M)$
    \If {not $S$}
        \State \textbf{break}
    \EndIf
    \State $S \gets \text{BothSideJudge}(p_{n-1}, p_n, M)$
    \If {not $S$}
        \State \textbf{break}
    \EndIf
    \State $n \gets n + 1$
\EndWhile
\State \Return $S$
\end{algorithmic}
\end{algorithm}

\vspace{-1mm}
\subsection{Multi-Mode Hybrid Trajectory Planning}
Obstacle density in real-world environments is highly heterogeneous, ranging from open areas with sparse obstructions to narrow passages with dense obstacles. Consequently, relying on a single, fixed trajectory optimization model is clearly insufficient to handle the diverse flight demands flexibly, often leading to suboptimal performance in both planning efficiency and trajectory quality. To address this issue, building upon the awareness of environmental complexity outlined in Sec. \ref{GEA}, this paper proposes a multi-mode hybrid intelligent planning mechanism. This mechanism incorporates three distinct planning modes: Fast Planning Mode (FPM), Standard Optimization Mode (SOM), and Emergency Obstacle Avoidance Mode (EAM). Each mode employs a trajectory optimization generation model with unique characteristics. Through the collaborative operation of these multiple models, leveraging their respective strengths while mitigating weaknesses, the mechanism achieves optimal planning efficiency. 

\begin{figure*}[!ht]
\centering
\begin{subfigure}[b]{0.48\textwidth}
    \centering
    \includegraphics[width=\linewidth]{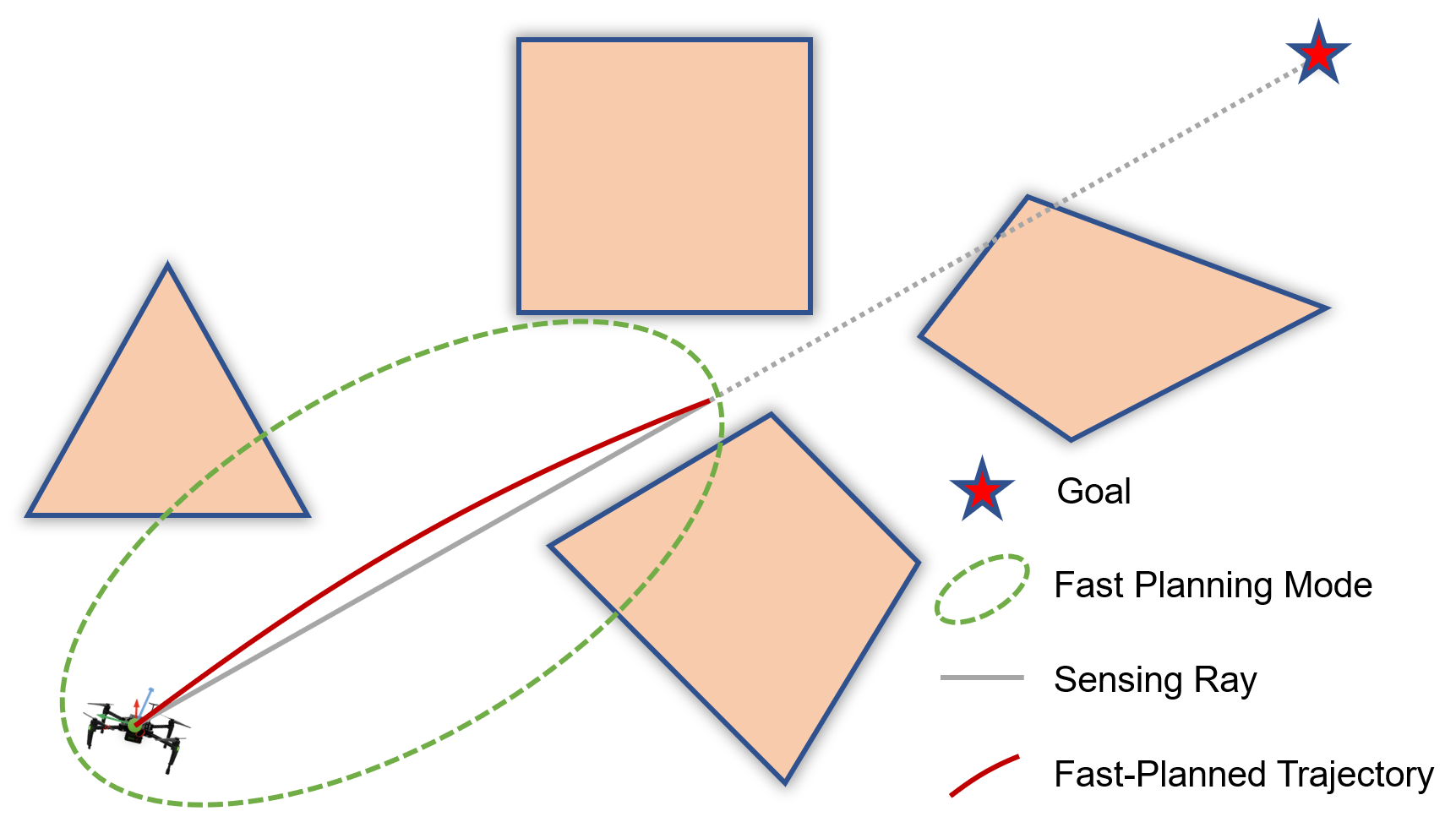} 
    \caption{Fast Planning Stage}
    \label{fig3a}
\end{subfigure}
\hfill
\begin{subfigure}[b]{0.48\textwidth}
    \centering
    \includegraphics[width=\linewidth]{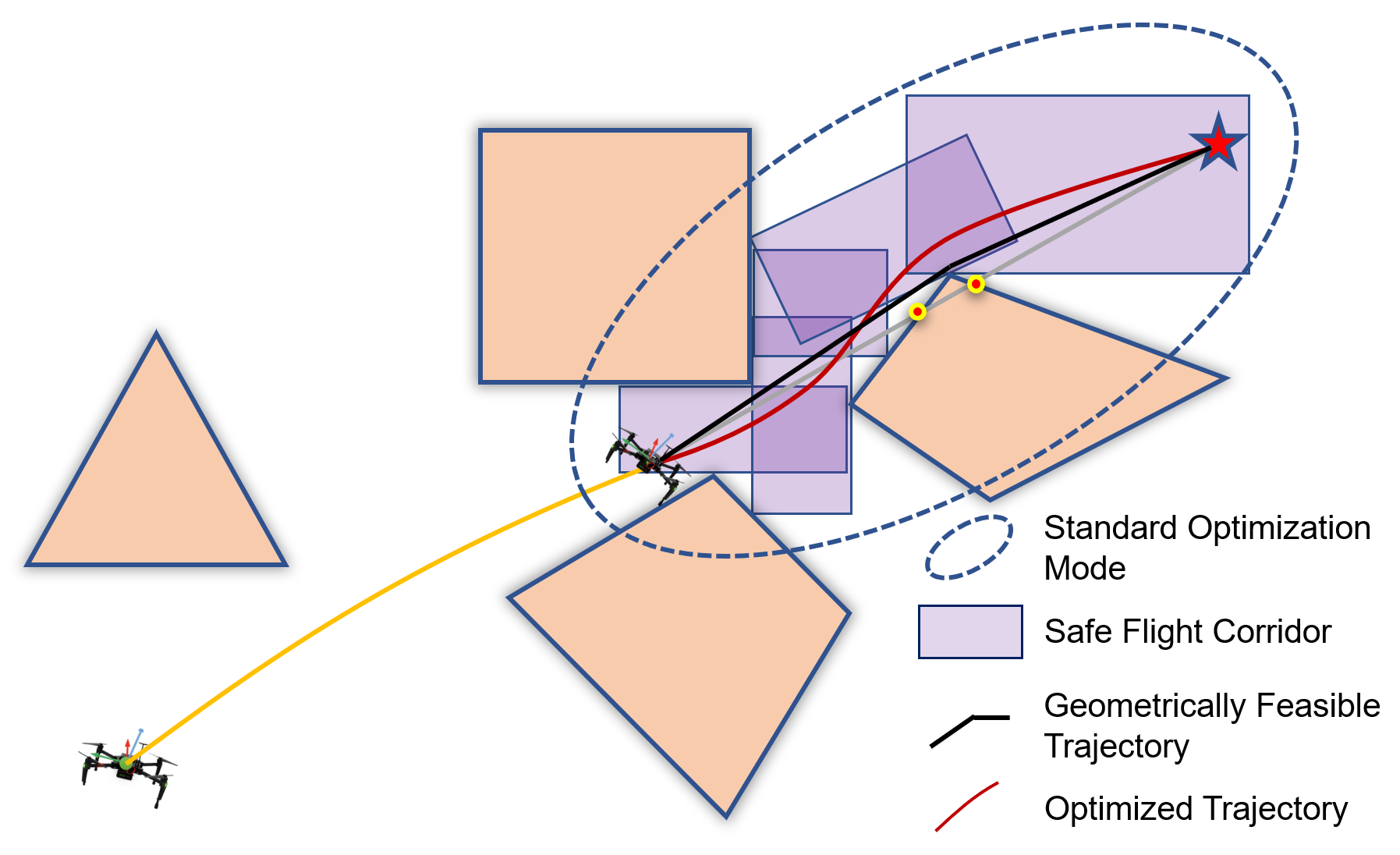} 
    \caption{Standard Optimization Stage}
    \label{fig3b}
\end{subfigure}
\\\vspace{3mm}
\begin{subfigure}[b]{0.8\textwidth}
    \centering
    \includegraphics[width=\linewidth]{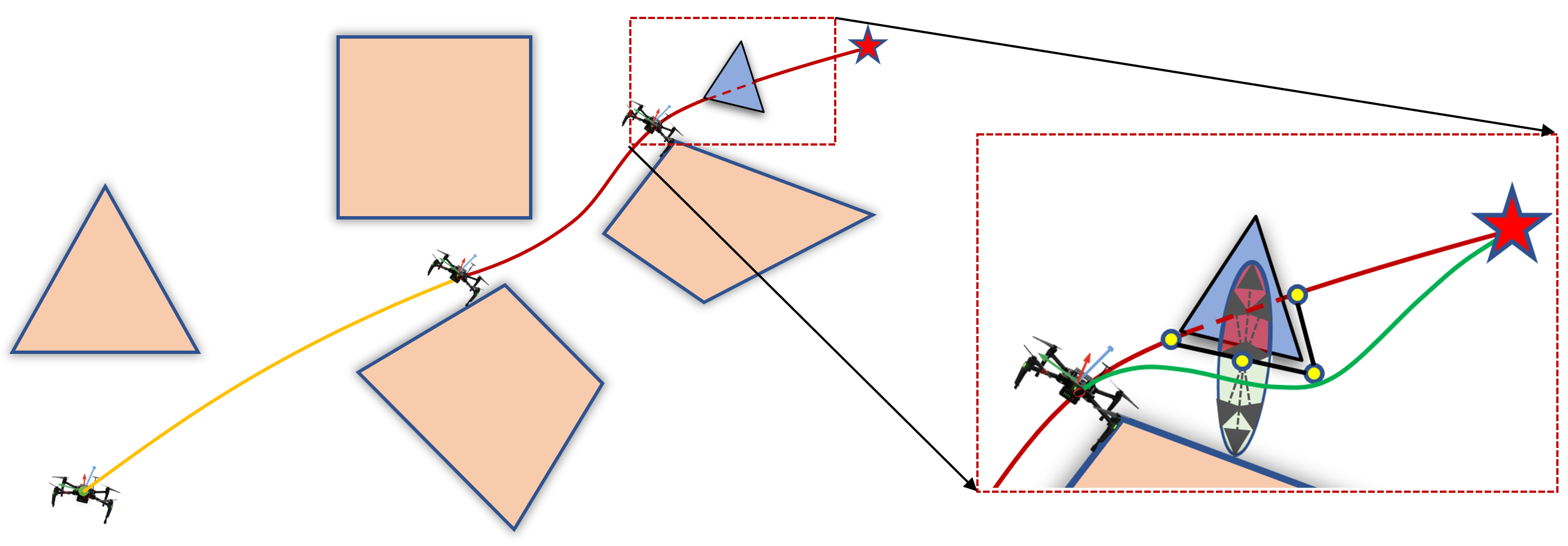} 
    \caption{Emergency Obstacle Avoidance Stage}
    \label{fig3c}
\end{subfigure}

\caption{Schematic diagram of the multi-mode hybrid planning mechanism. When no obstacles are detected by the sensing ray, as shown in (a), the system operates in fast planning mode to rapidly generate high-quality trajectories (red curves) without incorporating safety constraints. Conversely, when obstacles are detected (indicated by red collision points in (b)), the system switches to standard optimization mode. In this mode, a spatiotemporally optimized avoidance trajectory is generated within the SFC constructed based on an initial collision-free geometric path. Once newly emerged obstacles (blue solid triangles in (c)) interfere with the ongoing flight trajectory, the emergency obstacle avoidance mode is activated immediately, triggering local re-optimization to promptly compute a safe and dynamically feasible alternative trajectory (green curve).}
\label{fig3}
\vspace{-4mm}
\end{figure*}

\subsubsection{Fast Planning Mode (FPM)} {When the safety flag $S$ is True, indicating no obstacles are detected along the sensing ray, as shown in Fig. \ref{fig3a}, the environment is presumed to be sparsely cluttered with obstacles or to pose negligible threat to the ongoing flight mission. In such scenarios, we can disregard trajectory safety considerations and focus on enhancing trajectory quality and computational efficiency. Therefore, we adopt the more direct and efficient planning model, which can rapidly generate high-quality trajectory without considering safety constraint\cite{wang2020alternating}:
\begin{align}
\min_{\mathbf{D}_P, \mathbf{T}} &\; J(\mathbf{D}_P, \mathbf{T}) \tag{1} \\
\text{s.t. } &\; \left\| \mathbf{P}^{(n)}(t) \right\| \leq \sigma_n, \ 0 \leq t \leq \left\| \mathbf{T} \right\|_1 \tag{2} \\
&\; \mathbf{P} = \Phi(\mathbf{D}_P, \mathbf{T}),\ 1 \leq n \leq N \nonumber
\end{align}
where $\mathbf{D}_P$ and $\mathbf{T}$ denote the boundary conditions to be optimized and the trajectory time allocation, respectively. $\sigma_n$ represents the feasibility constraint, and $\left\| \mathbf{T} \right\|_1$ indicates the total trajectory duration. The trajectory $\mathbf{P}$ is generated by the mapping $\mathbf{P} = \Phi(\mathbf{D}_P, \mathbf{T})$. Throughout the optimization process, the $n$-th-order derivative of the trajectory is constrained not to exceed the threshold $\sigma_n$ (e.g., limits on velocity and acceleration) over the interval $0 \leq t \leq \left\| \mathbf{T} \right\|_1$, thereby ensuring dynamic feasibility. By disregarding safety considerations and building upon extensive validation, this method can efficiently compute a spatiotemporally optimal trajectory that satisfies all dynamic constraints.}

\subsubsection{Standard Optimization Mode (SOM)} {When $S$ is False, indicating the perception ray passes through an obstacle or a narrow space (as shown in Fig. \ref{fig3b}), the obstacle is deemed a safety threat to the current flight mission, necessitating standard obstacle avoidance trajectory planning. In such scenarios, we aim to rapidly generate a trajectory that strictly guarantees both safety and trajectory quality to achieve high-quality, safe drone flight. To achieve this, we first employ the A* algorithm to rapidly search for a locally safe and passable trajectory (black line segment in Fig. \ref{fig3b}) and construct SFC (purple region in Fig. \ref{fig3b}); Subsequently, based on the concept of hard constraints, we use the same method in \cite{wang2022geometrically} to construct multiple inequality constraints according to the solution space indicated by SFC, and then solve for the spatiotemporal optimal flight trajectory that satisfies all constraints. The overall optimization function to be adopted is as follows:
\begin{align}
\min_{p(t), T} &\; \int_0^T u(t)^T W u(t) dt + \rho(T) \tag{3} \\
\text{s.t. } &\; u(t) = p^{(s)}(t),\ \forall t \in [0,T], \nonumber \\
&\; \mathcal{G}\left(p(t), \dots, p^{(s)}(t)\right) \leq 0,\ \forall t \in [0,T], \nonumber \\
&\; p(t) \in \mathcal{F},\ \forall t \in [0,T], \nonumber \\
&\; p^{[s-1]}(0) = \bar{p}_o,\ p^{[s-1]}(T) = \bar{p}_f \tag{4}
\end{align}
where $p(t)$ denotes the flight trajectory. $T$ and $W$ represent the time and a positive diagonal matrix of the flight trajectory, respectively. $u(t)$ denotes the control input of the UAV. $\bar{p}_o$ and $\bar{p}_f$ denote the initial and final states of the flight trajectory, respectively. $\rho(T)$ is the time penalty term, $\mathcal{G}\left(p(t), \dots, p^{(s)}(t)\right)$ used to adjust the relative magnitude between control consumption and trajectory time consumption. By solving this optimization function through transformation, a spatiotemporally optimal flight trajectory can be rapidly generated within a safe flight corridor, as detailed in the referenced paper \cite{wang2022geometrically}. However, due to the limited solution space of hard constraint methods and their susceptibility to noise, this approach may fail to plan effectively in complex obstacle regions. In such cases, to ensure flight continuity, an emergency obstacle avoidance mode is triggered to adopt the more adaptable soft constraint method. }

\subsubsection{Emergency Avoidance Mode (EAM)} 
Since FPM produces result quickly but lack safety, while SOM offers high quality but suffer from relatively slow planning and weak adaptability, EAM will be swiftly triggered when newly detected obstacles threaten the current flight trajectory during operation, which ensures safety and planning success rates. In such scenarios, priority shifts from planning quality to trajectory real-time performance and safety. Compared to FPM that disregard safety, or SOM requiring time-consuming SFC construction, soft-constraint methods excel in both real-time capability and adaptability, making them the optimal choice. While soft-constraint approaches cannot guarantee strict compliance with predefined constraints, they offer superior adaptability. Additionally, it is crucial to note that in emergency situations, a certain degree of dynamic boundary violation is clearly acceptable to ensure flight safety and continuity. Therefore, for emergency obstacle avoidance, we adopt the core concept of EGO-Planner \cite{zhou2020ego}, known for its high computational efficiency and ESDF-free operation. However, unlike the standard EGO-Planner, which relies on a relatively fixed weighting scheme in its optimization, this paper introduces a more adaptive and robust optimization method. Specifically, while EGO-Planner leverages a locally searched geometric path (e.g., the black geometric path in Fig. \ref{fig3c}) generated by A* to replace ESDF for gradient evaluation, our method further analyzes this path using the narrow-space detection algorithm proposed in Sec. \ref{GEA}. The detection outcome then dynamically guides the adjustment of optimization weights. The overall objective function adopted is formulated as follows: 
\begin{table*}[!ht]
    \centering
    \begin{threeparttable}
    \footnotesize
    \caption{Performance comparison of four methods in different environments.}
    \label{tab:performance}
    \begin{tabular*}{\linewidth}{@{\extracolsep{\fill}} c l *{12}{r} } 
        \toprule
        \multirow{2}{*}{\textbf{Scene}} & \multirow{2}{*}{\textbf{Method}} & \multicolumn{4}{c}{\textbf{Flight time (s)}} & \multicolumn{4}{c}{\textbf{Flight distance (m)}} & \multicolumn{4}{c}{\textbf{Energy (m$^2$/s$^5$)}} \\
        \cline{3-14}
        & & \textbf{Avg} & \textbf{Std} & \textbf{Max} & \textbf{Min} & \textbf{Avg} & \textbf{Std} & \textbf{Max} & \textbf{Min} & \textbf{Avg} & \textbf{Std} & \textbf{Max} & \textbf{Min} \\
        \multirow{4}{*}{0} 
        & FastPlanner & \textbf{15.39} & 0.21 & \textbf{15.71} & \textbf{15.08} & 40.06 & 0.06 & 40.14 & 40.00 & 21.78 & 14.92 & 49.84 & 10.46 \\
        & EGOPlanner & 17.84 & \textbf{0.02} & 17.87 & 17.80 & 40.44 & 0.08 & 40.55 & 40.35 & 46.63 & 2.60 & 50.37 & 42.31 \\
        & ROTP & 16.22 & 0.11 & 16.44 & 16.13 & 40.02 & 0.15 & 40.31 & 39.91 & \textbf{9.42} & 0.89 & \textbf{10.25} & \textbf{7.73} \\
        & Proposed & 16.52 & \textbf{0.02} & 16.55 & 16.49 & \textbf{39.89} & \textbf{0.01} & \textbf{39.90} & \textbf{39.89} & 15.56 & \textbf{0.06} & 15.66 & 15.50 \\
        \hline
        \multirow{4}{*}{0.05} 
        & FastPlanner & 17.51 & 1.66 & 21.42 & 15.50 & 42.89 & 2.04 & 46.95 & 40.49 & 98.77 & 72.68 & 299.76 & 42.99 \\
        & EGOPlanner & 18.23 & 0.63 & 19.78 & 17.62 & 41.14 & 0.42 & 41.83 & 40.62 & 148.53 & 37.89 & 222.63 & 100.13 \\
        & ROTP & 17.29 & 0.37 & \textbf{17.88} & \textbf{16.64} & 40.64 & 0.63 & 41.94 & 40.13 & 67.71 & 31.97 & 127.51 & 28.13 \\
        & Proposed & \textbf{17.04} & \textbf{0.34} & 18.11 & 16.67 & \textbf{40.20} & \textbf{0.25} & \textbf{41.01} & \textbf{40.02} & \textbf{43.93} & \textbf{22.83} & \textbf{112.08} & \textbf{26.53} \\
        \hline
        \multirow{4}{*}{0.1} 
        & FastPlanner & 18.20 & 1.87 & 21.71 & \textbf{15.39} & 43.44 & 2.21 & 47.71 & 40.76 & 125.43 & 52.85 & 237.56 & \textbf{41.77} \\
        & EGOPlanner & 18.45 & \textbf{0.31} & \textbf{19.08} & 17.73 & 41.24 & \textbf{0.27} & \textbf{41.84} & 40.90 & 205.29 & 41.94 & 281.53 & 135.20 \\
        & ROTP & 18.69 & 0.95 & 21.36 & 17.58 & 41.87 & 0.95 & 43.65 & 40.20 & 106.33 & 41.93 & 174.33 & 39.58 \\
        & Proposed & \textbf{17.89} & 0.86 & 19.85 & 16.97 & \textbf{40.84} & 0.62 & 41.96 & \textbf{40.14} & \textbf{80.91} & \textbf{28.92} & \textbf{134.19} & 44.71 \\
        \hline
        \multirow{4}{*}{0.2} 
        & FastPlanner & 20.97 & 2.57 & 25.80 & \textbf{16.74} & 46.81 & 2.58 & 50.66 & 41.96 & 256.32 & 113.76 & 542.59 & 115.84 \\
        & EGOPlanner & 19.91 & 1.96 & 23.87 & 17.99 & 42.38 & 1.17 & 45.30 & 41.33 & 261.16 & 87.43 & 477.40 & 169.32 \\
        & ROTP & 19.23 & \textbf{0.80} & \textbf{21.29} & 18.10 & 42.97 & 1.76 & 46.99 & 40.56 & \textbf{129.92} & \textbf{38.24} & \textbf{195.80} & 57.08 \\
        & Proposed & \textbf{18.62} & 1.35 & 22.00 & 17.06 & \textbf{41.66} & \textbf{1.16} & \textbf{43.95} & \textbf{40.45} & 136.43 & 50.00 & 247.06 & \textbf{78.87} \\
        \bottomrule
    \end{tabular*}
    \tablenotes[para,flushleft]{\footnotesize *Bold values denote the optimal results for each scenario in the corresponding metrics.}
    \end{threeparttable}
    \vspace{-4mm}
\end{table*}
\begin{equation}
\begin{aligned}
\min_{Q} J &= \lambda_1(\mathcal{N}) J_s + \lambda_2(\mathcal{N}) J_c + \lambda_3 J_f \\
\lambda_1(\mathcal{N}) &= 
\begin{cases} 
\lambda_1^{\text{high}} & \text{if } \mathcal{N} = 0, \\
\lambda_1^{\text{low}} & \text{if } \mathcal{N} = 1
\end{cases}, \quad
\lambda_2(\mathcal{N}) = 
\begin{cases} 
\lambda_2^{\text{low}} & \text{if } \mathcal{N} = 0, \\
\lambda_2^{\text{high}} & \text{if } \mathcal{N} = 1
\end{cases}
\end{aligned}
\tag{5}
\end{equation}
where $Q$ denotes the control points of the B-spline. $\mathcal{N}$ represents the narrow space detection state: $\mathcal{N} = 1$ indicates a narrow passage area, while $\mathcal{N} = 0$ indicates an open passage area. $J_s$, $J_c$ and $J_f$ denote the penalty terms for trajectory smoothness, safety, and dynamic feasibility, respectively (specific calculation methods are provided in \cite{zhou2020ego}). 
$\lambda_1(\mathcal{N})$, $\lambda_2(\mathcal{N})$ and $\lambda_3$ are the penalty coefficients for the three penalty terms. The coefficient adjustment rule is:
1) In narrow passages ($\mathcal{N} = 1$): Adopt high-safety and low-smoothness weighting (decrease $\lambda_1$, increase $\lambda_2$).
2) In open passages ($\mathcal{N} = 0$): Adopt high-smoothness and low-safety weighting (increase $\lambda_1$, decrease $\lambda_2$).

This optimization function enables real-time, robust generation of safe, re-optimized local trajectories, improving the stability of feasible trajectory generation.
\vspace{-1mm}
\subsection{Lazy Replanning Strategy}
Due to the limited sensing range of sensors and the demand for high planning efficiency, trajectory planning typically operates within a small local area. Consequently, the results of a single trajectory planning session are insufficient to support the entire flight mission, and continuous trajectory replanning based on real-time environmental data is required to ensure the UAV safely reaches its target location. Current planning methods typically adopt fixed-high-frequency replanning strategies to ensure trajectory quality and safety. However, such strategies inevitably trigger numerous unnecessary replanings (no or limited improvement in terms of security, continuity, and optimality), wasting limited onboard computational resources. To address this issue, this paper proposes a lazy local replanning strategy: 
\subsubsection{Replanning range}
{To minimize unnecessary replanning in few-obstacles scenarios, each trajectory planning operation is conducted over a range exceeding the sensor's maximum sensing range, meaning portions of the planned trajectory will extend into unknown environment. While this approach cannot guarantee trajectory safety within unknown areas, it avoids triggering replanning when such areas contain no obstacles or when newly encountered obstacles do not impact the planned trajectory. Thus, this strategy reduces unnecessary planning while preserving flight trajectory quality and conserving computational resources. However, to avoid overconfidence and excessive planning time caused by overly long single-run distances, we typically limit the planning range to a distance achievable at maximum speed within 5 seconds of flight.}
\subsubsection{Replanning conditions}
{Local trajectory replanning is triggered only under two conditions: 1) Since current trajectory planning methods typically set the target point's velocity to zero, replanning is initiated when the current flight trajectory has been executed beyond the halfway point. This aims to maintain high-speed flight throughout the journey and prevent deceleration at the trajectory's end from affecting current flight performance. 2) Since portions of the trajectory occur in uncharted environments, local replanning will be rapidly triggered for emergency obstacle avoidance when newly detected obstacles threaten the safety of the ongoing flight trajectory. Through these strategies, we achieve autonomous obstacle avoidance for UAVs in complex environments using fewer computational resources, without compromising flight quality or safety.}

\begin{figure}[ht]
\centering
\begin{subfigure}[b]{0.5\textwidth}
    \centering
    \includegraphics[width=\linewidth]{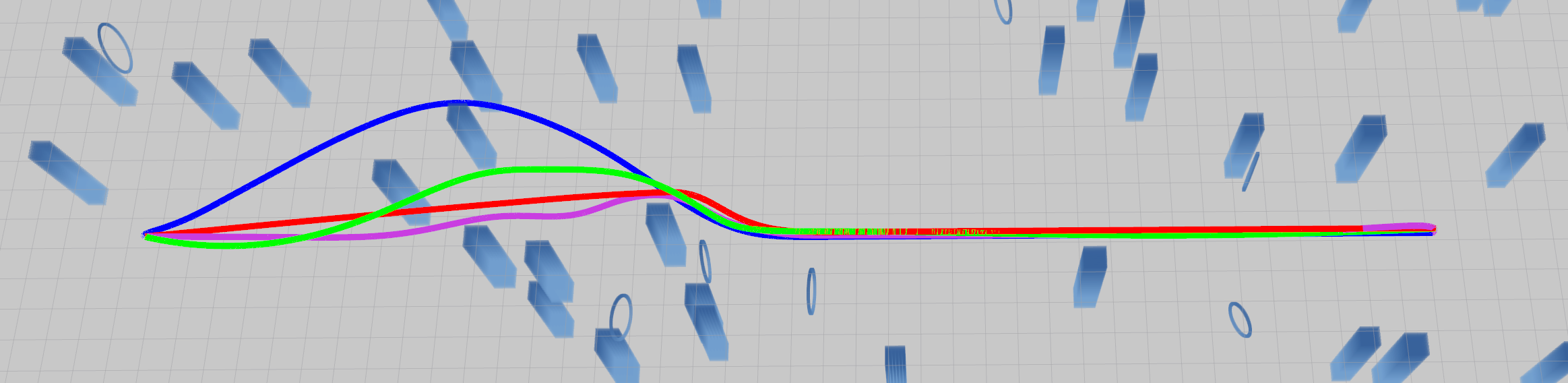}
    \caption{0.05 obs./m$^2$}
    \label{fig5a}
\end{subfigure}
\vspace{3mm}
\begin{subfigure}[b]{0.5\textwidth}
    \centering
    \vspace{3mm}
    \includegraphics[width=\linewidth]{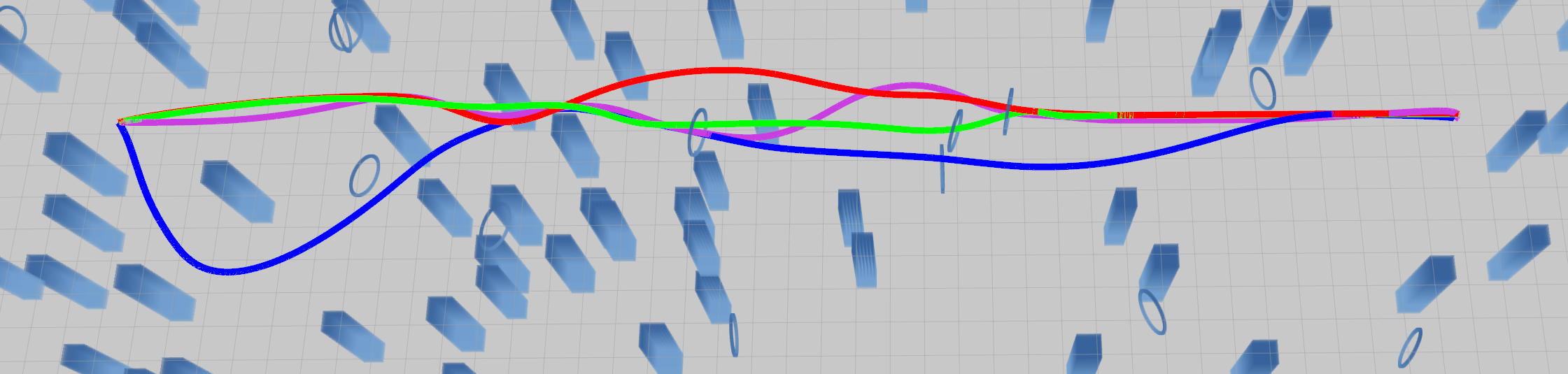}
    \caption{0.1 obs./m$^2$}
    \label{fig5b}
\end{subfigure}
\begin{subfigure}[b]{0.5\textwidth}
    \centering
    \includegraphics[width=\linewidth]{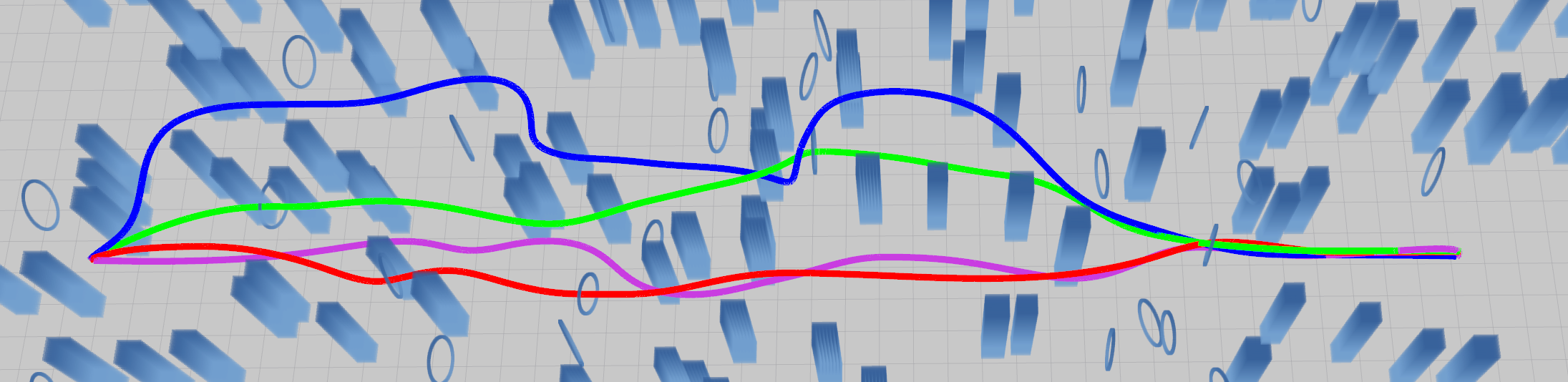}
    \caption{0.2 obs./m$^2$}
    \label{fig5c}
\end{subfigure}
\caption{Flight trajectory comparison of the four methods in different environments. The red, pink, green, and blue trajectories correspond to the proposed method, EGO-Planner, ROTP, and FastPlanner, respectively. It should be noted that for scenario with 0 obs./m$^2$ (i.e., an obstacle-free environment), the trajectories generated by all methods are straight lines and completely overlap. Therefore, comparative trajectory plots for this scenario are not displayed.}
\label{fig4.5}
\vspace{-4mm}
\end{figure}

\section{EXPERIMENTAL IMPLEMENTATION AND RESULTS}

\begin{figure*}[ht]
\centering
\begin{subfigure}[b]{0.48\textwidth}
    \centering
    \includegraphics[height=5cm,width=0.9\linewidth]{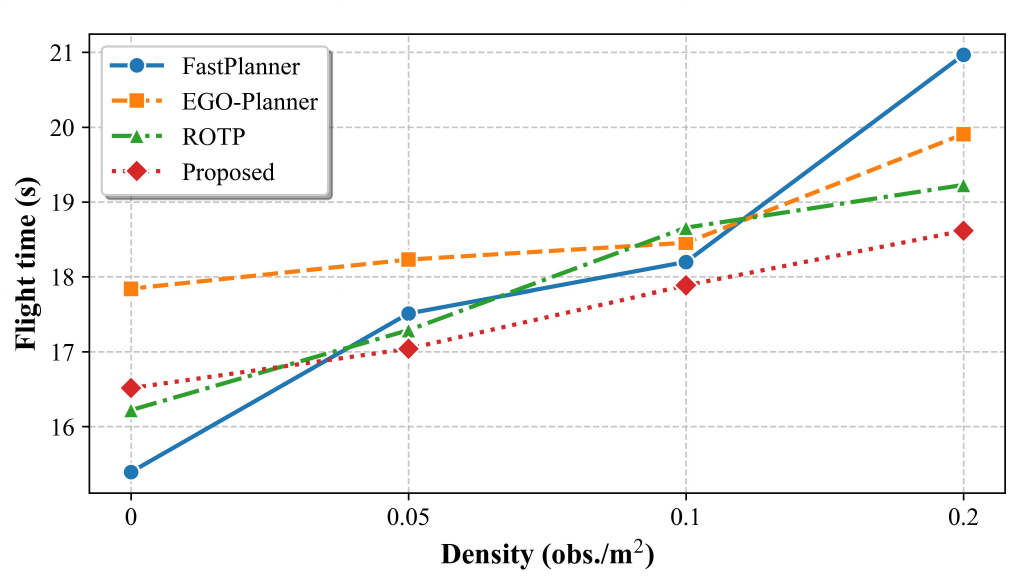}
    \caption{Flight time (s)}
    \label{fig5a}
\end{subfigure}
\hfill
\begin{subfigure}[b]{0.48\textwidth}
    \centering
    \includegraphics[height=5cm,width=0.9\linewidth]{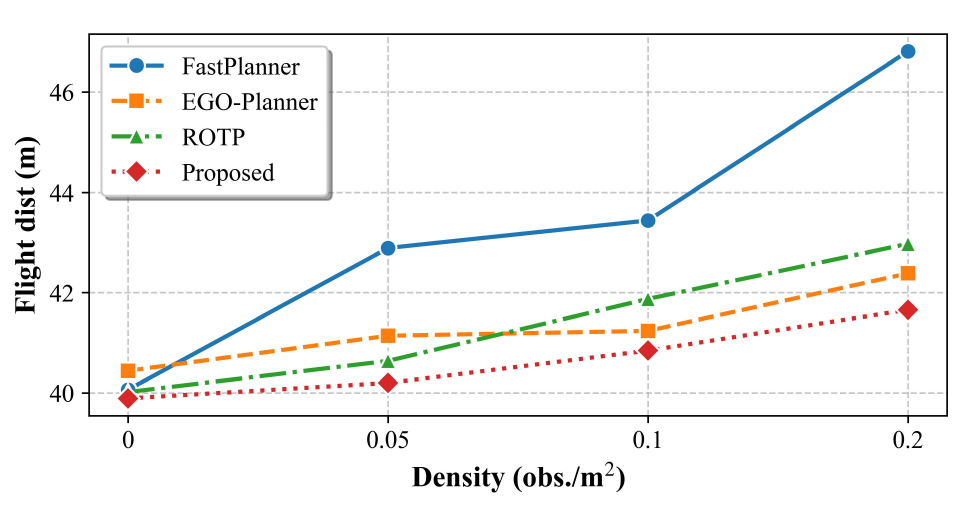}
    \caption{Flight dist (m)}
    \label{fig5b}
\end{subfigure}

\vspace{0.2cm}

\begin{subfigure}[b]{0.48\textwidth}
    \centering
    \includegraphics[height=5cm,width=0.9\linewidth]{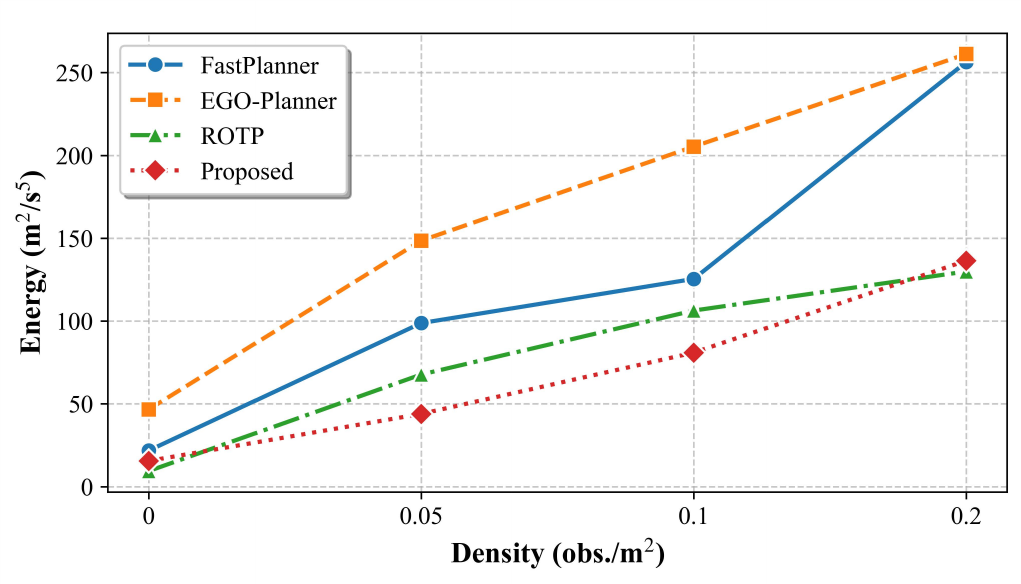}
    \caption{Energy}
    \label{fig5c}
\end{subfigure}
\hfill
\begin{subfigure}[b]{0.48\textwidth}
    \centering
    \includegraphics[height=5cm,width=0.9\linewidth]{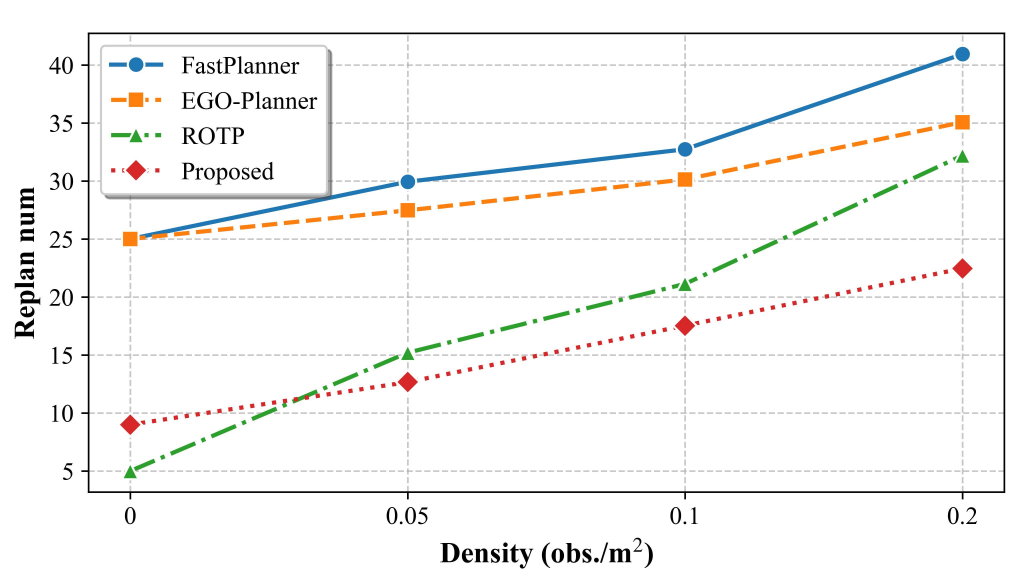}
    \caption{The number of planning iterations}
    \label{fig5d}
\end{subfigure}
\caption{Performance trends of the four planning methods in different environments: an evaluation of flight time, flight distance, flight energy, and the number of replanning iterations.}
\label{fig5}
\end{figure*}

\begin{table*}[htbp]
    \centering
    \begin{threeparttable}
    \caption{Comparison of the four methods in terms of planning iterations and average computational cost}
    \label{tab1:performance}
    \begin{tabularx}{\linewidth}{*{9}{X}}
        \toprule
         \multirow{2}{*}{\textbf{Scene}} & \multicolumn{4}{c}{\textbf{Avg. Number of Planning Iterations}} & \multicolumn{4}{c}{\textbf{Avg. Computational Cost Per Iteration (ms)}} \\
        \cline{2-9}
        & \textbf{FastP.} & \textbf{EgoP.} & \textbf{ROTP} & \textbf{Ours} & \textbf{FastP.} & \textbf{EgoP.} & \textbf{ROTP} & \textbf{Ours} \\
        0     & 25.0  & 25.0 & \textbf{5.0}  & 9.0  & 0.9  & 0.8  & 3.0  & \textbf{0.1}  \\
        0.05  & 29.9  & 27.5 & 15.2 & \textbf{12.7} & 2.0  & 1.1  & 2.5  & \textbf{0.3}  \\
        0.1   & 32.7  & 30.1 & 21.1 & \textbf{17.5} & 2.2  & 1.2  & 2.7  & \textbf{0.7}  \\
        0.2   & 40.9  & 35.1 & 32.2 & \textbf{22.5} & 2.7  & 1.4  & 2.6  & \textbf{0.9}  \\
        \bottomrule
    \end{tabularx}
    \tablenotes[para,flushleft]{\footnotesize *Bold values denote the optimal results for each scenario in the corresponding metrics.}
    \end{threeparttable}
    \vspace{-3mm}
\end{table*}

\subsection{Benchmark Comparison}
To comprehensively evaluate the performance of the proposed method, we conducted detailed comparative experiments in simulation environments against three state-of-the-art (SOTA) algorithms: FastPlanner\cite{zhou2019robust}, EGO-Planner\cite{zhou2020ego}, and ROTP\cite{zhao2023robust}. Both EGO-Planner and FastPlanner are well-known motion planning frameworks that have gained widespread recognition and adoption due to their excellent performance in trajectory quality and planning efficiency. ROTP is a previous work from our research team, excels not only in local trajectory replanning but also maintains stable navigation capability in environments with large obstacles. All algorithms in our simulations were implemented using their open-source codes. The maximum velocity and acceleration of the UAV dynamics were set to 3.0 m/s and 2.0 m/s$^2$, respectively, while the sensor field of view (FOV) and maximum sensing range were configured as [80\textdegree, 60\textdegree] and 4.5 m. All experiments were performed on a hardware platform with an Intel Core i9-13980HX processor (2.20 GHz), 32 GB of RAM, Ubuntu 20.04 and ROS Noetic.

Obstacle density in real-world environments is highly heterogeneous, encompassing both sparsely obstructed regions and densely cluttered areas. To accurately simulate this variability and rigorously evaluate algorithm performance across different scenarios, we generated random maps in a $50 \times 50 \times 3$ m$^3$ volume with four distinct obstacle density levels: 0 obs./m$^2$ (obstacle-free environment), 0.05 obs./m$^2$, 0.1 obs./m$^2$, and 0.2 obs./m$^2$. For each density, three distinct maps were randomly created. Subsequently, all planning methods were evaluated through five flight trials per map using same start and goal positions.  

The detailed experimental results are summarized in TABLE \ref{tab:performance}. The experimental results demonstrate that across four distinct obstacle density environments, the proposed algorithm achieved the optimal performance. It delivered the best or second-best results among the four methods in terms of flight time, flight distance, energy consumption, and the standard deviation, which demonstrates that the proposed method not only exhibits outstanding planning efficiency but also possesses excellent stability. Fig. \ref{fig5} illustrates the trends of the aforementioned metrics as obstacle density increases. It is evident that our approach exhibits a trend closer to linear variation. This not only demonstrates the enhanced robustness of our method but also validates the efficacy of our multi-mode hybrid replanning mechanism. It shows that the framework seamlessly adapts its planning mode to mitigate the challenges posed by rising obstacle density, thereby ensuring graceful performance scaling. As shown in Fig. \ref{fig4.5}, we also present a comparison of flight trajectories across different obstacle density environments. The flight trajectories of our method were also more rational and smoother, consistent with the benchmark results. Notably, as shown in TABLE \ref{tab1:performance}, compared to the other three methods, our approach significantly outperforms them in terms of average number of planning iterations and computational cost per iteration, substantially reducing computational overhead. This experimental outcome primarily stems from our algorithm's more adaptable planning methodology and replanning strategy. Our method dynamically selects the most appropriate planning model based on the awareness of the real-time scene. In open areas without safety risks, it employs the FPM for fast trajectory generation, prioritizing planning efficiency to reduce planning time. In obstacle-rich avoidance scenarios, it utilizes the SOM's hard-constraint planning method to generate high-quality obstacle-avoidance trajectories, enhancing trajectory quality. When newly emerging obstacles threaten trajectory safety, EAM performs real-time local trajectory re-optimization, enhancing the urgency, safety, and continuity of obstacle avoidance. Furthermore, by adopting a long-range lazy replanning strategy instead of fixed high-frequency replanning, new trajectory planning is triggered only when necessary (speed or safety deteriorates). This significantly reduces the number of planning iterations while ensuring flight safety and efficiency, further lowering computational resource consumption.

In contrast, FastPlanner\cite{zhou2019robust} employs control space sampling for initial trajectory search and utilizes heuristic functions for efficiency. Therefore, as shown in Fig. \ref{fig4.5}, although its backend optimization improves the trajectory quality in smoothness and safety, it still suffers from reduced trajectory optimality, prone to extensive detours and getting stuck in local optima. The EGO-Planner\cite{zhou2020ego}, overly reliant on the initial guidance trajectory, prioritizes returning to the original path after each obstacle avoidance. This causes the trajectory to oscillate around the initial path, increasing both planning time and energy consumption. Although the ROTP\cite{zhao2023robust} also adopts a planning strategy combining hard and soft constraints, it predominantly utilizes the hard-constraint-based algorithm named GCOPTER\cite{wang2022geometrically}. While this method generates extremely high-quality trajectories, its delayed replanning strategy and excessive reliance on the result of GCOPTER often lead to an overemphasis on trajectory smoothness, resulting in elongated paths.

\subsection{Real-World Experiment}
\begin{figure}[!ht]
\centering
\includegraphics[width=0.9\linewidth]{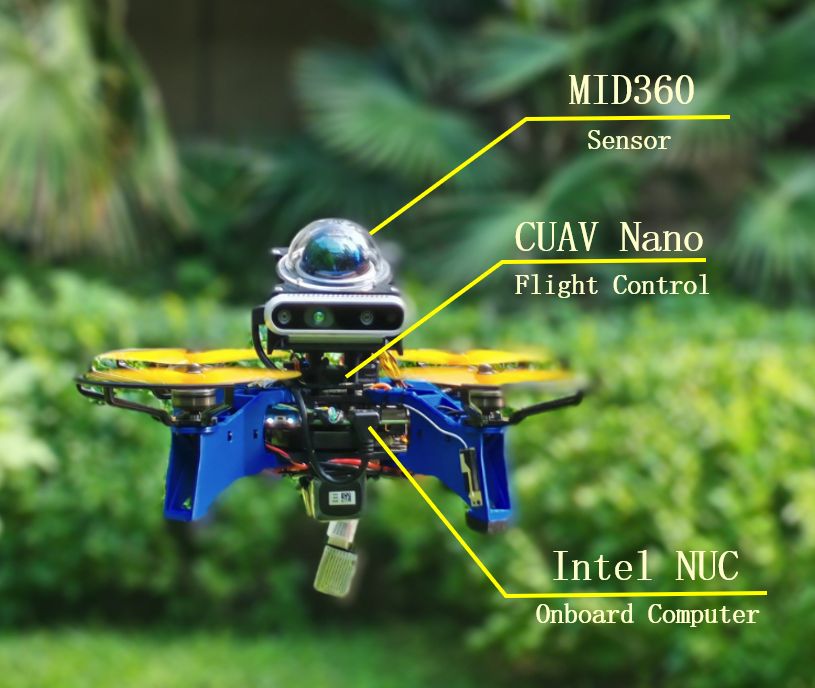}
\caption{The fully self-developed intelligent quadrotor (250mm wheelbase) used in real-world experiments.}
\label{fig7}
\vspace{-3mm}
\end{figure}
To validate the effectiveness of the proposed method, we conducted practical flight experiment in outdoor environment using a fully self-developed intelligent UAV platform. The UAV and its configuration are shown in Fig. \ref{fig7}, equipped with a MID360 LiDAR for obstacle data acquisition. During the experiment, we utilized  \cite{chen2022direct} to provide the state estimates. All algorithm modules ran on an onboard computer featuring an Intel Core i7-13980HX (2.20 GHz) processor with 16GB of memory, operating under Ubuntu 20.04 with the ROS Neotic software environment.

\begin{figure*}[!ht]
\centering
\includegraphics[width=\linewidth]{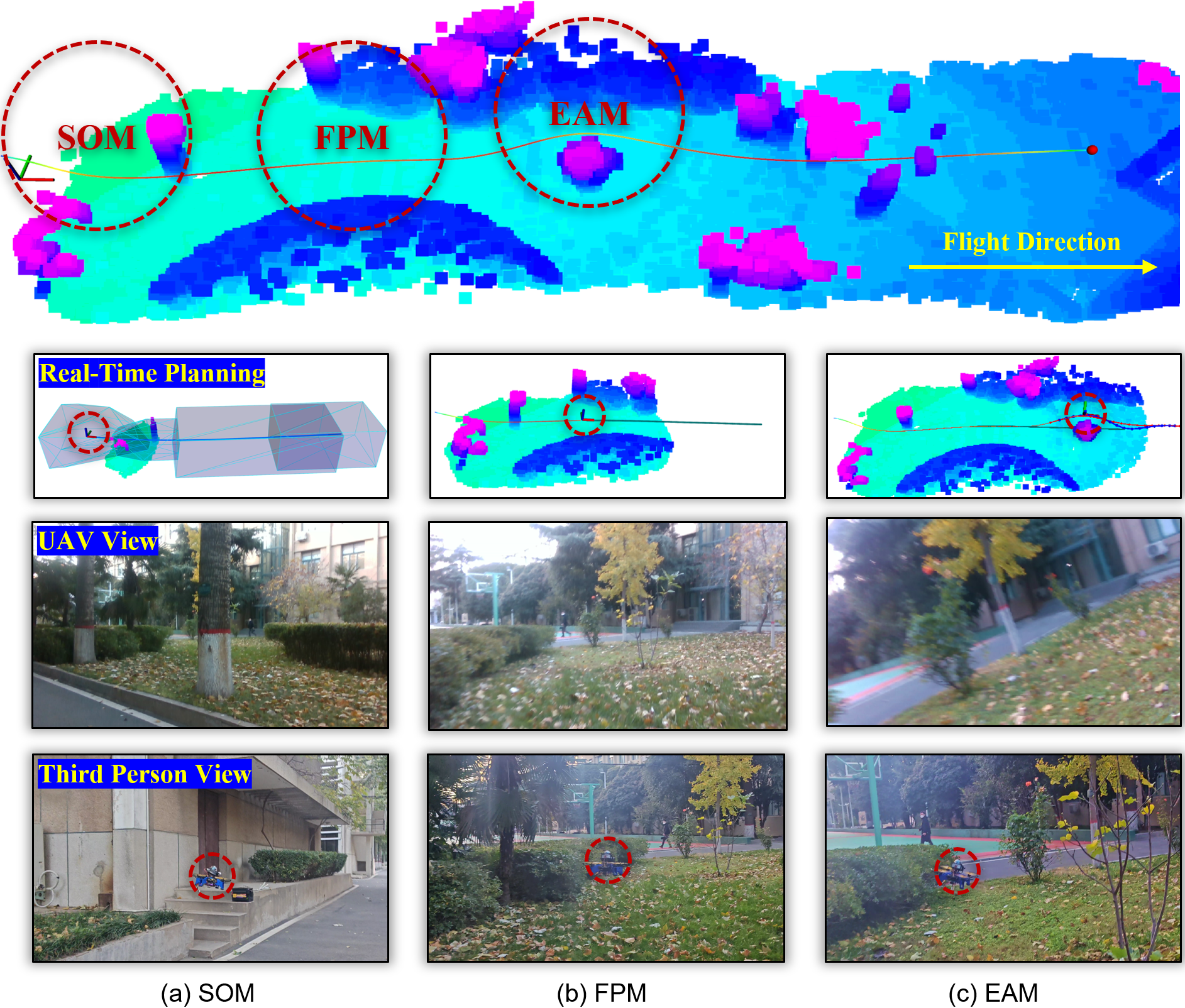}
\caption{Experimental results of obstacle avoidance flight in real-world outdoor environment by integrating the proposed method with the customized quadrotor, which includes the complete flight trajectory, the concurrently constructed environmental map, and the status when different modes (SOM, FPM, EAM) were triggered (as shown in red dashed circle). In (a), the blue transparent polyhedrons are the constructed SFC, while the blue trajectory is the optimal flight path generated by SOM. In (b), the cyan trajectory is the fast planning result by FPM. In (c), the red trajectory is locally re-optimized result by EAM.}
\label{fig8}
\vspace{-4mm}
\end{figure*}

The experimental scenario features multiple plant obstacles of varying sizes, representing a typical unstructured and irregular environment. Within this environment, the UAV is tasked with flying to a location 31 meters from the starting point. During this flight mission, the drone must perform multiple planning iterations to avoid obstacles, effectively validating the effectiveness of the algorithm proposed in this paper. Additionally, for this flight experiment, we set the maximum flight speed and acceleration limits for the UAV to 3.0 m/s and 2.0 m/s$^2$, respectively, with the sensor sensing range set to 5 m. Ultimately, the UAV completed the flight in 13.1 seconds, covering a trajectory length of 32.3 meters. It achieved a maximum speed of 3.04 m/s and an average speed of 2.46 m/s. Additionally, the entire flight mission consumed 11 replanning iterations, with each iteration averaging 1.3 ms. The complete flight trajectory, constructed obstacle map, multi-view images, and detailed planning information in different stages are shown in Fig. \ref{fig8}. Meanwhile, the UAV's flight trajectory is color-coded to represent its instantaneous speed, using a gradient from blue to red, where increasing redness indicates higher velocity. From the experimental results, it can be clearly observed that through the collaboration of different planning modes (SOM, FPM, and EAM), the UAV successfully and rapidly completed the obstacle avoidance task, with a smooth and rational overall flight trajectory. The UAV maintained high-speed flight for most of the mission, except during the initial, final, and emergency avoidance phases. This experiment verifies that the proposed algorithm effectively supports safe, fast, and efficient navigation for UAVs in unknown environments. 

\section{Conclusion}
In this paper, we propose a multi-mode hybrid replanning method, capable of autonomously selecting the optimal planning model for trajectory generation according to the real-time spatial awareness. First, we design a goal-oriented spatial safety assessment method to rapidly evaluate the impact of obstacles within the upcoming navigation area on the current flight mission and to provide prior guidance for subsequent trajectory planning. Second, we develop a hybrid intelligent planning mechanism incorporating three distinct modes (FPM, SOM, and EAM) tailored to the characteristics of different planning models. This mechanism selects the optimal trajectory optimization model for path generation based on the navigation spatial awareness, leveraging the strengths of each planning model to achieve a more reasonable trade-off between planning efficiency and trajectory quality while ensuring flight safety. Finally, a lazy replanning strategy was developed to trigger trajectory replanning only when necessary, significantly reducing the waste of onboard computational resources. Comprehensive comparative experiments evaluated the algorithm's performance, and practical flight tests were conducted using an fully self-developed intelligent UAV hardware platform in real-world environments. The experimental results robustly validate the effectiveness and superiority of the proposed algorithm, demonstrating its capability to support fast, safe, and autonomous UAV flight.


\bibliographystyle{Bibliography/IEEEtranTIE}
\bibliography{Bibliography/IEEEabrv,Bibliography/BIB_xx-TIE-xxxx}\ 

@article{du2025efficient,
  title={An efficient uav coverage path planning method for 3-d structures},
  author={Du, Jiaxin and Huang, Baoqi and Jia, Bing},
  journal={IEEE Internet of Things Journal},
  year={2025},
  publisher={IEEE}
}

@article{liu2024dawn,
  title={DAWN: Dynamic task planning of multi-UAV with two-layer optimization mechanism in uncertain environments},
  author={Liu, Daqian and Fei, Bowen and Bao, Weidong and Zhu, Xiaomin and Li, Xiaoqing},
  journal={IEEE Internet of Things Journal},
  volume={11},
  number={23},
  pages={37813--37830},
  year={2024},
  publisher={IEEE}
}

@article{guo2025global,
  title={A Global-Local Collaborative and Decomposition-Based Multi-Objective Evolutionary Optimization Method for UAV 3-D Path Planning},
  author={Guo, Jianchong and Wan, Yuting and Ma, Ailong and Zhong, Yanfei},
  journal={IEEE Internet of Things Journal},
  year={2025},
  publisher={IEEE}
}

@article{zhang2025hybrid,
  title={Hybrid Deep Reinforcement Learning for UAV Inspection in Large-Scale Wind Farms: Deployment and Routing Optimization},
  author={Zhang, Xiao-Yu and Yu, Huiming and Zheng, Xingnan and Wang, Hui and Mu, Chaoxu and Guo, Peiqian},
  journal={IEEE Transactions on Industrial Informatics},
  year={2025},
  publisher={IEEE}
}

@article{zhou2020ego,
  title={Ego-planner: An esdf-free gradient-based local planner for quadrotors},
  author={Zhou, Xin and Wang, Zhepei and Ye, Hongkai and Xu, Chao and Gao, Fei},
  journal={IEEE Robotics and Automation Letters},
  volume={6},
  number={2},
  pages={478--485},
  year={2020},
  publisher={IEEE}
}

@article{zhou2023racer,
  title={Racer: Rapid collaborative exploration with a decentralized multi-uav system},
  author={Zhou, Boyu and Xu, Hao and Shen, Shaojie},
  journal={IEEE Transactions on Robotics},
  volume={39},
  number={3},
  pages={1816--1835},
  year={2023},
  publisher={IEEE}
}

@article{ye2020tgk,
  title={Tgk-planner: An efficient topology guided kinodynamic planner for autonomous quadrotors},
  author={Ye, Hongkai and Zhou, Xin and Wang, Zhepei and Xu, Chao and Chu, Jian and Gao, Fei},
  journal={IEEE Robotics and Automation Letters},
  volume={6},
  number={2},
  pages={494--501},
  year={2020},
  publisher={IEEE}
}

@inproceedings{quan2021eva,
  title={EVA-planner: Environmental adaptive quadrotor planning},
  author={Quan, Lun and Zhang, Zhiwei and Zhong, Xingguang and Xu, Chao and Gao, Fei},
  booktitle={2021 IEEE International Conference on Robotics and Automation (ICRA)},
  pages={398--404},
  year={2021},
  organization={IEEE}
}

@article{dominguez2025cooperative,
  title={Cooperative robotic exploration of a planetary skylight surface and lava cave},
  author={Dom{\'\i}nguez, Ra{\'u}l and P{\'e}rez-del-Pulgar, Carlos and Paz-Delgado, Gonzalo J and Polisano, Fabio and Babel, Jonathan and Germa, Thierry and Dragomir, Iulia and Ciarletti, Val{\'e}rie and Berthet, Anne-Claire and Danter, Leon Cedric and others},
  journal={Science Robotics},
  volume={10},
  number={105},
  pages={eadj9699},
  year={2025},
  publisher={American Association for the Advancement of Science}
}

@article{wang2025unlocking,
  title={Unlocking aerobatic potential of quadcopters: Autonomous freestyle flight generation and execution},
  author={Wang, Mingyang and Wang, Qianhao and Wang, Ze and Gao, Yuman and Wang, Jingping and Cui, Can and Li, Yuan and Ding, Ziming and Wang, Kaiwei and Xu, Chao and others},
  journal={Science Robotics},
  volume={10},
  number={101},
  pages={eadp9905},
  year={2025},
  publisher={American Association for the Advancement of Science}
}

@article{zhou2026hfch,
  title={HFCH: Hybrid frontier guided fast UAV autonomous exploration for complete and high-quality mapping in unknown environment},
  author={Zhou, Yuquan and Yan, Li and Han, Yaxi and Zhu, Longze and Zhao, Yinghao and Xie, Hong},
  journal={Advanced Engineering Informatics},
  volume={69},
  pages={103834},
  year={2026},
  publisher={Elsevier}
}

@article{zhao2023robust,
  title={Robust planning system for fast autonomous flight in complex unknown environment using sparse directed frontier points},
  author={Zhao, Yinghao and Yan, Li and Dai, Jicheng and Hu, Xiao and Wei, Pengcheng and Xie, Hong},
  journal={Drones},
  volume={7},
  number={3},
  pages={219},
  year={2023},
  publisher={MDPI}
}

@article{chen2022direct,
  title={Direct lidar-inertial odometry: Lightweight lio with continuous-time motion correction},
  author={Chen, Kenny and Nemiroff, Ryan and Lopez, Brett T},
  journal={arXiv preprint arXiv:2203.03749},
  year={2022}
}

@inproceedings{richter2016polynomial,
  title={Polynomial trajectory planning for aggressive quadrotor flight in dense indoor environments},
  author={Richter, Charles and Bry, Adam and Roy, Nicholas},
  booktitle={Robotics Research: The 16th International Symposium ISRR},
  pages={649--666},
  year={2016},
  organization={Springer}
}

@article{wang2020alternating,
  title={Alternating minimization based trajectory generation for quadrotor aggressive flight},
  author={Wang, Zhepei and Zhou, Xin and Xu, Chao and Chu, Jian and Gao, Fei},
  journal={IEEE Robotics and Automation Letters},
  volume={5},
  number={3},
  pages={4836--4843},
  year={2020},
  publisher={IEEE}
}

@inproceedings{usenko2017real,
  title={Real-time trajectory replanning for MAVs using uniform B-splines and a 3D circular buffer},
  author={Usenko, Vladyslav and Von Stumberg, Lukas and Pangercic, Andrej and Cremers, Daniel},
  booktitle={2017 IEEE/RSJ International Conference on Intelligent Robots and Systems (IROS)},
  pages={215--222},
  year={2017},
  organization={IEEE}
}

@inproceedings{ratliff2009chomp,
  title={CHOMP: Gradient optimization techniques for efficient motion planning},
  author={Ratliff, Nathan and Zucker, Matt and Bagnell, J Andrew and Srinivasa, Siddhartha},
  booktitle={2009 IEEE international conference on robotics and automation},
  pages={489--494},
  year={2009},
  organization={IEEE}
}

@article{zhou2022swarm,
  title={Swarm of micro flying robots in the wild},
  author={Zhou, Xin and Wen, Xiangyong and Wang, Zhepei and Gao, Yuman and Li, Haojia and Wang, Qianhao and Yang, Tiankai and Lu, Haojian and Cao, Yanjun and Xu, Chao and others},
  journal={Science Robotics},
  volume={7},
  number={66},
  pages={eabm5954},
  year={2022},
  publisher={American Association for the Advancement of Science}
}

@inproceedings{mellinger2011minimum,
  title={Minimum snap trajectory generation and control for quadrotors},
  author={Mellinger, Daniel and Kumar, Vijay},
  booktitle={2011 IEEE international conference on robotics and automation},
  pages={2520--2525},
  year={2011},
  organization={IEEE}
}

@article{wang2022geometrically,
  title={Geometrically constrained trajectory optimization for multicopters},
  author={Wang, Zhepei and Zhou, Xin and Xu, Chao and Gao, Fei},
  journal={IEEE Transactions on Robotics},
  volume={38},
  number={5},
  pages={3259--3278},
  year={2022},
  publisher={IEEE}
}

@article{ren2025safety,
  title={Safety-assured high-speed navigation for MAVs},
  author={Ren, Yunfan and Zhu, Fangcheng and Lu, Guozheng and Cai, Yixi and Yin, Longji and Kong, Fanze and Lin, Jiarong and Chen, Nan and Zhang, Fu},
  journal={Science Robotics},
  volume={10},
  number={98},
  pages={eado6187},
  year={2025},
  publisher={American Association for the Advancement of Science}
}

@inproceedings{ren2022bubble,
  title={Bubble planner: Planning high-speed smooth quadrotor trajectories using receding corridors},
  author={Ren, Yunfan and Zhu, Fangcheng and Liu, Wenyi and Wang, Zhepei and Lin, Yi and Gao, Fei and Zhang, Fu},
  booktitle={2022 IEEE/RSJ International Conference on Intelligent Robots and Systems (IROS)},
  pages={6332--6339},
  year={2022},
  organization={IEEE}
}

@article{tordesillas2021mader,
  title={MADER: Trajectory planner in multiagent and dynamic environments},
  author={Tordesillas, Jesus and How, Jonathan P},
  journal={IEEE Transactions on Robotics},
  volume={38},
  number={1},
  pages={463--476},
  year={2021},
  publisher={IEEE}
}

@article{zhao2023autonomous,
  title={Autonomous exploration method for fast unknown environment mapping by using UAV equipped with limited FOV sensor},
  author={Zhao, Yinghao and Yan, Li and Xie, Hong and Dai, Jicheng and Wei, Pengcheng},
  journal={IEEE Transactions on Industrial Electronics},
  volume={71},
  number={5},
  pages={4933--4943},
  year={2023},
  publisher={IEEE}
}

@article{zhou2019robust,
  title={Robust and efficient quadrotor trajectory generation for fast autonomous flight},
  author={Zhou, Boyu and Gao, Fei and Wang, Luqi and Liu, Chuhao and Shen, Shaojie},
  journal={IEEE Robotics and Automation Letters},
  volume={4},
  number={4},
  pages={3529--3536},
  year={2019},
  publisher={IEEE}
}

@article{zhao2021robust,
  title={Robust and efficient trajectory replanning based on guiding path for quadrotor fast autonomous flight},
  author={Zhao, Yinghao and Yan, Li and Chen, Yu and Dai, Jicheng and Liu, Yuxuan},
  journal={Remote Sensing},
  volume={13},
  number={5},
  pages={972},
  year={2021},
  publisher={Multidisciplinary Digital Publishing Institute}
}

@inproceedings{tordesillas2019faster,
  title={Faster: Fast and safe trajectory planner for flights in unknown environments},
  author={Tordesillas, Jesus and Lopez, Brett T and How, Jonathan P},
  booktitle={2019 IEEE/RSJ international conference on intelligent robots and systems (IROS)},
  pages={1934--1940},
  year={2019},
  organization={IEEE}
}

@article{zhou2021raptor,
  title={Raptor: Robust and perception-aware trajectory replanning for quadrotor fast flight},
  author={Zhou, Boyu and Pan, Jie and Gao, Fei and Shen, Shaojie},
  journal={IEEE Transactions on Robotics},
  volume={37},
  number={6},
  pages={1992--2009},
  year={2021},
  publisher={IEEE}
}

\end{document}